%% file: main.tex
\documentclass[sn-aps]{sn-jnl}

\usepackage{graphicx}%
\usepackage{multirow}%
\usepackage{amsmath,amssymb,amsfonts}%
\usepackage{amsthm}%
\usepackage{mathrsfs}%
\usepackage[title]{appendix}%
\usepackage{xcolor}%
\usepackage{textcomp}%
\usepackage{manyfoot}%
\usepackage{booktabs}%
\usepackage{algorithm}%
\usepackage{algorithmicx}%
\usepackage{algpseudocode}%
\usepackage{listings}%
\usepackage{algpseudocode,algorithm}
\usepackage{soul}
\usepackage{dsfont}
\usepackage{tabularx}
\usepackage{fancyhdr}
\fancypagestyle{firstpage}{
  \fancyhf{} 
  \rfoot{} 
  \cfoot{} 
  \lfoot{Published in \textit{Quality and Reliability Engineering International} (2023). \href{https://doi.org/10.1002/qre.3392}{https://doi.org/10.1002/qre.3392}} 
}

\begin{document}

\title[Article Title]{Robust online active learning}


\author*[1,2]{\fnm{Davide} \sur{Cacciarelli}}\email{dcac@dtu.dk}

\author[1,3]{\fnm{Murat} \sur{Kulahci}}\email{muku@dtu.dk}

\author[2]{\fnm{John Sølve} \sur{Tyssedal}}\email{john.tyssedal@ntnu.no }

\affil[1]{\orgdiv{Department of Applied Mathematics and Computer Science}, \orgname{Technical University of Denmark}, \orgaddress{\city{Kgs. Lyngby}, \country{Denmark}}}

\affil[2]{\orgdiv{Department of Mathematical Sciences}, \orgname{Norwegian University of Science and Technology}, \orgaddress{\city{Trondheim}, \country{Norway}}}

\affil[3]{\orgdiv{Department of Business Administration, Technology and Social Sciences}, \orgname{Luleå University of Technology}, \orgaddress{\city{Luleå}, \country{Sweden}}}


\abstract{\input{0_abstract}}





\maketitle
\thispagestyle{firstpage} 


\section{Introduction} \label{sec:introduction}
\input{1_introduction}

\section{Background and related work} \label{sec:preliminaries}
\input{2_background}

\section{Methods} \label{sec:methods}
\input{3_methods}

\section{Experiments} \label{sec:experiments}
\input{4_experiments}

\section{Discussion} \label{sec:disc}
\input{5_discussion}

\section{Conclusion} \label{sec:conclusion}
\input{6_conclusion}


\begin{appendices}

\input{7_appendix}

\end{appendices}

\typeout{}
\bibliography{sn-bibliography}

\end{document}

%% file: 0_abstract.tex
In many industrial applications, obtaining labeled observations is not straightforward as it often requires the intervention of human experts or the use of expensive testing equipment. In these circumstances, active learning can be highly beneficial in suggesting the most informative data points to be used when fitting a model. Reducing the number of observations needed for model development alleviates both the computational burden required for training and the operational expenses related to labeling. Online active learning, in particular, is useful in high-volume production processes where the decision about the acquisition of the label for a data point needs to be taken within an extremely short time frame. However, despite the recent efforts to develop online active learning strategies, the behavior of these methods in the presence of outliers has not been thoroughly examined. In this work, we investigate the performance of online active linear regression in contaminated data streams. Our study shows that the currently available query strategies are prone to sample outliers, whose inclusion in the training set eventually degrades the predictive performance of the models. To address this issue, we propose a solution that bounds the search area of a conditional D-optimal algorithm and uses a robust estimator. Our approach strikes a balance between exploring unseen regions of the input space and protecting against outliers. Through numerical simulations, we show that the proposed method is effective in improving the performance of online active learning in the presence of outliers, thus expanding the potential applications of this powerful tool.

%% file: 1_introduction.tex
Predictive models often need to be trained on a large amount of labeled data before being deployed. However, in industrial applications data is often abundant only in an unlabeled form. Active learning strategies provide a solution to this problem by prioritizing the labeling of the most useful instances for building the model, thus accelerating the convergence of its learning curve \cite{Kumar2020}. Active learning problems can be classified into three macro-scenarios \cite{Settles2009}. The first and most studied scenario is the pool-based scenario, where the learner can select the most useful instances to be labeled by maximizing an evaluation criterion over a closed set of observations. The second scenario is referred to as membership query synthesis, and it allows the learner to query the labels of synthetically generated instances rather than those sampled from the process distribution. Finally, the third scenario is online, or stream-based, active learning \cite{OALSurvey}. In this case, the unlabeled observations are drawn sequentially by the learner, which must immediately decide whether to keep the instance and query its label or discard it. While many researchers have been working on active learning in the recent years, the pool-based scenario has received the most attention \cite{Chan2018,Ge2014}. 
Although online active learning has become more popular in the last few years \cite{Liu2015,Bouguelia2016,Lughofer2012,Shan2019,Krawczyk2017}, the majority of the methods have been developed for classification tasks \cite{Lughofer2017}. An interesting approach to online active learning for fuzzy regression models has been proposed by Lughofer \cite{Lughofer2018}. Other researchers tried to adapt the optimality criteria of the experimental design theory to the online active linear regression framework \cite{Riquelme2017multi, Riquelme2017AAAI,Fontaine2021,SBAL}. Linear regression models are still very useful in industrial applications as they can be efficiently trained on a small number of observations. They are able to offer a straightforward interpretation, along with the possibility of constructing confidence intervals on the parameter estimates \cite{AlvarezMelis,Efron2004}. They can also be easily coupled with variable selection and robust estimation methods. Furthermore, whereas many pool-based active learning approaches employ ensemble methods or complex models, linear models can support online active learning due to the decreased computational cost associated with model training and updating.

\begin{figure}[h]
  \centering
  \includegraphics[width=\linewidth]{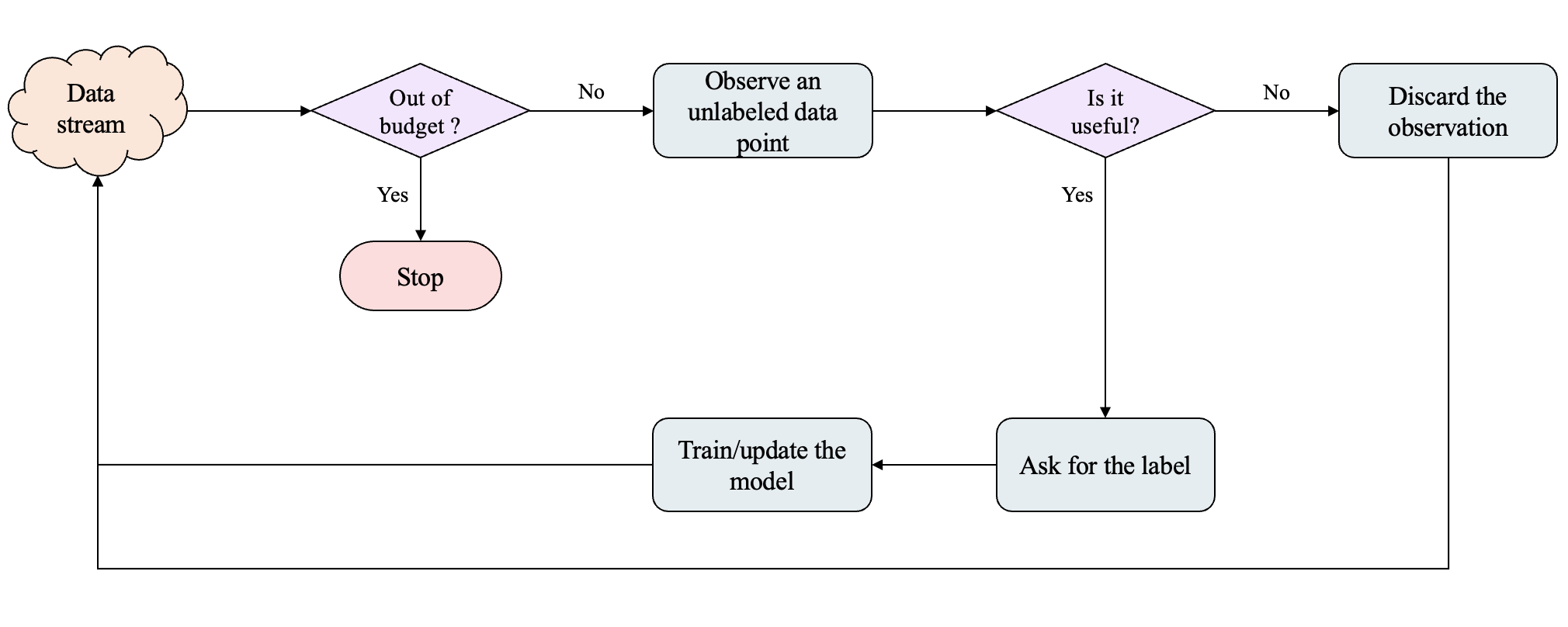}
  \caption{General online active learning flowchart..}
  \label{fig:flowchart}
\end{figure}

Figure \ref{fig:flowchart} depicts a general online active learning flowchart. The main difference among the query strategies lies in how they assess the usefulness of an unlabeled instance when the learner samples it from the data stream. Another important aspect is the assumptions on the input distribution. Indeed, despite the increased interest in the online active linear regression framework, the performance of the sampling strategies in the presence of outliers has not been thoroughly explored. The few works we are aware of that analyze this issue, are related to the pool-based scenario. Deldossi et al. \cite{Deldossi2022} highlighted how sampling methods based on D-optimality are affected by outliers and high leverage points. Zhao et al. \cite{Zhao2022} focused on robust active representations based on the $\ell_{2,p}$-norm constraints for selecting highly representative data. Finally, He et al. \cite{He2017} emphasized the problem of being prone to sample outliers while proposing a semi-supervised active learning strategy for multivariate time series classification, using uncertainty and local density. 

In this paper, we study the problem of learning from contaminated data streams with limited sampling resources. We first investigate the effects of outliers on the sampling decisions made by state-of-the-art online active learning approaches for linear regression, and successively propose a solution for this issue. It should be noted that the presence of outliers considered in this work cannot be tackled using traditional anomaly detection methods. Indeed, most unsupervised anomaly detection strategies rely on the assumption that a large training set free from outliers, usually referred to as phase I data in the statistical process control literature, is available beforehand \cite{OAE,Nguyen2019,Zhou2017,Ruff2021}.  However, this assumption is violated in many practical applications \cite{Qiu2022}, especially in label-scarce scenarios where few to no labels are available before the beginning of the active learning routine. The proposed strategy for online active learning utilizes a double-threshold approach to limit the search area of a conditional D-optimality algorithm (CDO). By using two thresholds, the strategy aims to identify informative data points while excluding outliers. In cases of highly contaminated environments, robust estimators based on the Huber and Tukey bisquare loss are employed.

The remainder of this paper is organized as follows. In Section \ref{sec:preliminaries}, we introduce the terminology and describe the sampling strategies that are used as the baseline in our analysis. Section \ref{sec:methods} offers a review on the use of robust estimators and introduces ways of modifying the CDO algorithm. In Section \ref{sec:experiments}, we test our approach using numerical simulations in four scenarios, using different contamination ratios. Section \ref{sec:disc} offers a discussion on the results obtained. Finally, Section \ref{sec:conclusion} provides some conclusions.

%% file: 2_background.tex
The labeled observations that are collected from the contaminated data stream are used to fit a linear model of the form

\begin{equation}
    \mathbf{y}=\mathbf{X} \boldsymbol{\beta} + \boldsymbol{\varepsilon}
\end{equation}

\noindent
where $\mathbf{y}$ is an $n \times 1$ vector of response variables, $\mathbf{X}$ is an $n \times p$ model matrix, $\boldsymbol{\beta}$ is a $p \times 1$ vector of regression coefficients, and $\boldsymbol{\varepsilon}$ is an $n \times 1$ vector representing the zero-mean Gaussian noise. Here, $n$ represents the total number of observations, and $p$ the number of variables. Before starting the active learning routine and the collection of additional labels, we assume to have at our disposal an initial set of labeled observations as in \cite{Burbidge2007,Ge2014,Ge2016}. This set is used to obtain an initial estimate $\widehat{\boldsymbol{\beta}}$ for the coefficients $\boldsymbol{\beta}$. Using an ordinary least squares (OLS) estimator, we have that $\widehat{\boldsymbol{\beta}}=\left(\mathbf{X}^\top \mathbf{X}\right)^{-1} \mathbf{X}^\top \mathbf{y}$. Then, the fitted linear regression model is $\widehat{\mathbf{y}}=\mathbf{X} \widehat{\boldsymbol{\beta}}$, and the residuals are obtained as $\mathbf{e}=\mathbf{y}-\widehat{\mathbf{y}}$. When the variables are highly correlated, a pre-whitening might be performed to avoid an ill-conditioned problem when computing $\left(\mathbf{X}^\top \mathbf{X}\right)^{-1}$. It should be noted that the matrix $\mathbf{X}^\top \mathbf{X}$ is important to obtain information about the design geometry. In particular, for a design composed of $n$ runs, the moment matrix, $\mathbf{ M }=(\mathbf{X}^\top \mathbf{X})⁄n$, plays a central role in the definition of optimal experimental designs. The two most commonly employed optimality criteria, which have been adapted for the online active learning scenario, are A-optimality and D-optimality. An A-optimal design is achieved by minimizing the trace of the inverse of the moment matrix $\mathbf{ M }$. It can be shown how this corresponds to minimizing the individual variances of the estimated coefficients. This approach has been adapted for the online active linear regression framework by Riquelme et al. \cite{Riquelme2017AAAI}. They proposed a norm-thresholding algorithm that only selects observations $\mathbf{ x }$ with large, scaled norm by estimating a threshold $\Gamma$ as

\begin{equation}
    \mathds{P}(\|\mathbf{x}\| \geq \Gamma)=\alpha
\end{equation}

\noindent
where $\alpha$ is the ratio of observations we are willing to label out of the incoming data stream. The probability distribution of the norms can be approximated using kernel density estimation (KDE) on a set of unlabeled observations $\mathbf{C}$, which can be regarded as a warm-up or calibration set and can either be retrieved from historical data or by observing the data stream for a while. Using this thresholding approach, we would be sampling, with high probability, observations that help achieve A-optimality. Given $n$ statistics, $(s_1, \ldots,s_n)$, KDE can be used to estimate the shape of an unknown distribution $f$ using

\begin{equation}
    \widehat{f}(s)=\frac{1}{n} \sum_{i=1}^n \frac{1}{h} K\left(\frac{s-s_i}{h}\right)
\end{equation}

\noindent
where the bandwidth $h$ is a positive number that is used to control the amount of smoothing, and the kernel $K$ is a smooth function such that $K(s) \geq 0, \int K(s) d s=1, \int s K(s) d s=0 \text { and } \sigma_K^2 \equiv \int s^2 K(s) d s>0$. In this paper, the Gaussian (Normal) kernel, $K(s)=(2 \pi)^{-1 / 2} e^{-s^2 / 2}$ is used.

D-optimality is another fundamental criterion \cite{John1975}, which takes both the variances and covariances of the model coefficients into account by maximizing the determinant of the moment matrix M. As in the case of A-Optimality, D-Optimality has been adapted to the online active learning scenario with the proposal of a conditional D-optimality (CDO) algorithm \cite{SBAL}. CDO suggests setting a threshold $\Gamma$ by using

\begin{equation}
\mathds{P}\left(\mathbf{x}_{l+1}^\top\left(\mathbf{X}_l^\top \mathbf{X}_l\right)^{-1} \mathbf{x}_{l+1} \geq \Gamma\right)=\alpha
\end{equation}

\noindent
where $\mathbf{X}_l$ is the model matrix with the $l$ labeled observations currently available and $\mathbf{x}_{l+1}$ is the unlabeled data point that is under evaluation. It can be shown that by selecting observations that maximize $\mathbf{x}_{l+1}^\top\left(\mathbf{X}_l^\top \mathbf{X}_l\right)^{-1} \mathbf{x}_{l+1}$, we are at the same time seeking D-optimality and labeling observations with a large unscaled prediction variance (UPV) \cite{Myers2016}, which is generally defined as

\begin{equation}
    \operatorname{UPV}(\mathbf{x})=\mathbf{x}^{(m) \top}\left(\mathbf{X}^\top \mathbf{X}\right)^{-1} \mathbf{x}^{(m)}
\end{equation}

\noindent
where $\mathbf{x}^{(m)}$ represents the data point where the UPV is being estimated, expanded to the model form (e.g., if polynomial features are added to the model). To estimate the threshold $\Gamma$, we use KDE after computing the UPV of all the observations in $\mathbf{C}$. The CDO intuition is coherent with the idea that a point for which we have a large UPV value represents a less explored region of the input space and will help, with high probability, attaining D-optimality, conditional on the already collected observations. The equivalence between sampling data points with high UPV and D-optimality is demonstrated in \cite{SBAL}.
Given these preliminaries, we now propose methods that are robust to the presence of outliers in the data stream.

%% file: 3_methods.tex
When training a linear regression model on a dataset corrupted by the presence of outliers, a simple yet effective solution is to resort to the use of robust estimators. An extensive overview of robust regression has been provided by Fox and Weisberg \cite{Fox2013}. In general, robust estimation methods attempt to estimate the coefficients $\widehat{\boldsymbol{\beta}}$ by minimizing a particular loss function given by

\begin{equation}
    \mathcal{J}=\sum_{i=1}^n \rho\left(e_i\right)=\sum_{i=1}^n \rho\left(y_i-\mathbf{x}_i \widehat{\boldsymbol{\beta}}\right)
\end{equation}

\noindent
where $\rho$ is a function that regulates the contribution of each residual to the loss, and $e_i$ is the residual for the $i$th observation $\left(\mathbf{x}_i, y_i\right)$. The function $\rho$ is nonnegative, equal to zero when the argument is zero, symmetrical and monotone in $|e|$. In the case of an OLS estimator, the loss is given by

\begin{equation}
    \rho_{OLS}=e^2
\end{equation}

\noindent
It can be seen how the objective function minimized by an OLS estimator is equally affected by all the observations for which we measure the residuals. Instead, robust estimators try to reduce the impact of observation with very large residuals on the estimation of $\widehat{\boldsymbol{\beta}}$. One of the most popular robust loss functions is the Huber loss \cite{Huber1964}, which is defined as

\begin{equation}
    \rho_H=\left\{\begin{array}{lll}
e^2 & \text { for }|e| \leq k \\
2 k|e|-k^2 & \text { for }|e|>k
\end{array}\right.
\end{equation}

\noindent
where $k$ is a tuning parameter, which is usually set to $1.345\sigma$ to achieve 95\% efficiency when the errors are normally distributed while keeping good protection against outliers \cite{Fox2013}. It can be seen how the contribution of each observation is reduced based on the magnitude of the corresponding residual. However, despite being much more robust than the OLS estimator, the Huber loss is still proportional to the magnitude of the residuals even when the absolute errors are larger than $k$. Conversely, the Tukey bisquare loss function \cite{Beaton1974} sets a threshold for the residuals, above which the value of the residuals does not influence the loss. The Tukey loss function is given by

\begin{equation}
    \rho_T= \begin{cases}\frac{k^2}{6}\left\{1-\left[1-\left(\frac{e}{k}\right)^2\right]^3\right\} & \text { for }|e| \leq k \\ \frac{k^2}{6} & \text { for }|e|>k\end{cases}
\end{equation}

\noindent
where the value of the tuning constant k is usually set up to $4.685\sigma$ \cite{Fox2013}. Besides using a Huber or Tukey loss to obtain a robust estimator, we consider the possibility of filtering out outliers while selecting the most informative observations from the data stream. To this extent, we propose an adaptation of the CDO algorithm, where instead of estimating a threshold, we define a bounded area of interest for the unscaled prediction variance of an observation as

\begin{equation} \label{eq:bcdo}
    \mathds{P}\left(\Gamma_1 \leq \mathbf{x}_{l+1}^\top\left(\mathbf{X}_l^\top \mathbf{X}_l\right)^{-1} \mathbf{x}_{l+1} \leq \Gamma_2\right)=\alpha
\end{equation}

\noindent
This approach is hereinafter referred to as bounded CDO. The idea is coherent with the method proposed by Hoaglin and Welsch \cite{Hoaglin1978,Chatterjee1986} of considering as potential outliers observations for which $\mathbf{x}_i^\top\left(\mathbf{X}^\top \mathbf{X}\right)^{-1} \mathbf{x}_i \geq 2 p / n$ is verified. The filtering approach suggested by Hoaglin and Welsch is also used by Deldossi et al. \cite{Deldossi2022}, in the offline scenario. Here, instead of opting for a fixed value for $\Gamma_2$, we use KDE with a Gaussian kernel to estimate $\Gamma_1$ and $\Gamma_2$. The upper limit $\Gamma_2$ is selected by determining a cut-off value $c$, which is related to the amount of protection against outliers that we would like to achieve. This value is a tuning constant similar to the $k$ used by robust estimators and, when possible, should be selected by exploiting previous knowledge of the process. Given the cut-off value c and the sampling rate $\alpha$, $\Gamma_2$ is given by the $100(1-c)\%$ percentile, and $\Gamma_1$ by the $100(1-c-\alpha)\%$ percentile. As anticipated in Section \ref{sec:preliminaries}, the threshold estimation is based on a set of unlabeled data, which is also used to estimate the covariance matrix $\boldsymbol{\Sigma}$ and whitening the observations to remove dependencies and facilitate the estimation of $\widehat{\boldsymbol{\beta}}$. At this stage, semi-supervised methods might also be considered to perform tasks like feature extraction and exploit all the information available in the unlabeled data \cite{He2017,Fernandes2020,Leng2013,Frumosu2018,Cacciarelli2022}.

\begin{algorithm}[h]
\caption{Bounded CDO}\label{alg:1}
\begin{algorithmic}[h]
\Require data stream $\textbf{S}$, initial random design $\textbf{X}$, warm-up length $m$, budget $B$
\State Set $\textbf{C}=\varnothing$ \Comment{Calibration set to estimate $\boldsymbol{\Sigma}$, $\Gamma_1$, $\Gamma_2$}
\State $i\gets 1$ \Comment{Timestamp}
\State $b\gets 0$ \Comment{Labeling cost}
\While{$i\leq m$} \Comment{Warm-up}
\State Observe the $i$th data point $x_i\in \textbf{S}$
\State Select $x_i$: $\textbf{C}=\textbf{C}\cup \textbf{x}_i$
\State $i\gets i+1$
\EndWhile
\State Estimate covariance matrix $\boldsymbol{\Sigma}$ from $\textbf{C}$ and perform eigendecomposition $\boldsymbol{\Sigma}=\textbf{U}\boldsymbol{\Lambda}\textbf{U}^\top$
\State Whiten the initial design by computing $\textbf{Z}=\boldsymbol{\Lambda}^{-1/2}\textbf{U}^\top\textbf{X}$
\State Whiten the calibration set by computing $\textbf{V}=\boldsymbol{\Lambda}^{-1/2}\textbf{U}^\top\textbf{C}$
\State Estimate $\Gamma_1$, $\Gamma_2$ by estimating the UPV of the model trained on $\textbf{Z}$ on the points in $\textbf{V}$
\While{$b\leq B$ and $i\leq |\textbf{S}|$}
\State Observe the $i$th data point $\textbf{x}_i\in \textbf{S}$
\State Whiten $\textbf{x}_i$ by computing $\textbf{z}_i=\boldsymbol{\Lambda}^{-1/2}\textbf{U}^\top\textbf{x}_i$
\If{$\Gamma_1\leq \textbf{z}_i^\top(\textbf{Z}^\top\textbf{Z})^{-1}\textbf{z}_i\leq \Gamma_2$}
\State Ask for the label $y_i$ and augment the labeled dataset: $\textbf{Z}=\textbf{Z}\cup \{\textbf{z}_i\}$
\State $b\gets b+1$ \Comment{Pay for the label}
\State Update thresholds $\Gamma_1$, $\Gamma_2$ using the augmented design
\Else
\State Discard $\textbf{x}_i$
\EndIf
\State $i\gets i+1$
\EndWhile
\State \textbf{return} $\textbf{Z}$
\end{algorithmic}
\end{algorithm}

Algorithm \ref{alg:1} provides a detailed explanation of how to implement the bounded CDO strategy for online active learning in a fixed-budget setting. The strategy involves collecting new labels and incorporating them into the design until a specified budget constraint B is reached. In some cases, it might be beneficial to anticipate the stop of the active learning routine if the marginal improvement of the model is no longer significant \cite{hitting}. Previous studies have proposed various stopping criteria to enhance the efficiency of data collection schemes based on active learning \cite{Zhang2017SC,Ishibashi2020,Ghayoomi2010,Laws2008,Zhu2008SC}. Appendix \ref{app:stop} explores how some of these approaches could be adapted to the regression framework. From a computational standpoint, the update of $\widehat{\boldsymbol{\beta}}$ is done by means of a complete retraining each time a new labeled example is added to the design. However, if the data matrix becomes considerably large and the time required for model updates increases, one may opt to update the model and estimate new thresholds when a batch of new observations is collected, aligning with the principles of batch-mode active learning \cite{Ren2022}. Additionally, incremental and recursive updating techniques can also be considered for improving computational efficiency.

The estimation of the UPV can be modified by taking into account the weight matrix obtained from the robust estimators. The weighted UPV ($\text{UPV}_w$) is estimated as follows

\begin{equation}
    \text{UPV}_{w}(\mathbf{x})=\mathbf{x}^{(m) \top}\left(\mathbf{X}^\top \mathbf{W} \mathbf{X}\right)^{-1} \mathbf{x}^{(m)}
\end{equation}

\noindent
where $\mathbf{W}$ represents the weight matrix used to downweigh the influence of outliers in the estimation of the regression parameters \cite{Fox2013}. Each element of the weight matrix $\mathbf{W}$ is a positive number that determines the weight given to each observation in the regression analysis. Larger weights correspond to observations with less outlier-like behavior, while smaller weights correspond to observations with more outlier-like behavior. The weight matrix $\mathbf{W}$ is a diagonal matrix, where each diagonal element corresponds to the weight assigned to a particular observation. In the case of an OLS estimator, we have $\mathbf{W}=\mathbf{I}_k$, as the weight given to each observation is not sensitive to the residual. In other words, $w_{OLS} (e)=1$, regardless of the specific residual observed. With a Huber estimator, $w_H (e)=1$ if $|e|\leq k$ and $w_H (e)= k/|e|$ if $|e|>k$. Finally, with a Tukey model, $w_T (e)=0$ if $|e|>k$ and to $w_T(e)=\left[1-(e / k)^2\right]^2$ if $|e|\leq k$. Then, to select the most informative observations while seeking protection against outliers, instead of estimating a single threshold, we define a bounded area of interest for the UPV of an observation as follows

\begin{equation}
    \mathds{P}\left(\Gamma_1 \leq \mathbf{x}_{l+1}^\top\left(\mathbf{X}_l^\top \mathbf{W} \mathbf{X}_l\right)^{-1} \mathbf{x}_{l+1} \leq \Gamma_2\right)=\alpha
\end{equation}

%% file: 4_experiments.tex
In the experiments, we evaluate the performance of the active learning strategies in four scenarios, according to the percentage of outliers affecting the data stream. We compare the bounded CDO strategy, coupled with OLS and robust estimators, to the norm-thresholding approach, standard CDO, and random sampling. When using random sampling, each time a new sample arrives, a number $r \sim \mathcal{U}(0,1)$ is generated and the data point is only selected if $r \geq 1-\alpha$,where $\alpha$ represents the labeling or sampling rate. The sampling strategies based on the use of robust estimators select the most informative data points using the standard UPV, as in Equation \ref{eq:bcdo}. The results obtained with the weighted prediction variance, $\text{UPV}_w$, were very similar and are included in the Appendix \ref{app:upv} for completeness. All the approaches receive as input the same random design and then they iteratively collect labeled observations until the budget constraint $B$ is met. The number of observations contained in the initial design is equal to $p+2$, where p is the number of process variables. We analyzed both the case of the initial design being outliers-free and contaminated. The results assuming the presence of outliers also in the initial design are included in the Appendix \ref{app:contaminated}. For each simulated scenario, the $i$th observation for the process variables, here considered a row vector, is generated according to a joint multivariate normal distribution

\begin{equation}
    \mathbf{x}_i \sim \mathcal{N}_p\left(\mathbf{0}, \mathbf{\Sigma}_0\right)
\end{equation}

\noindent
where $\mathbf{\Sigma}_0$ is given by $\sigma_{\mathbf{x}}^2 \mathbf{I}$. The corresponding response is obtained using

\begin{equation}
    y_i=\mathbf{x}_i \boldsymbol{\beta}+\varepsilon_i, \quad \text { where } \varepsilon_i \sim \mathcal{N}\left(0, \sigma_{\varepsilon}^2\right)
\end{equation}

\noindent
For normal data points, we used $\sigma_{\mathbf{x}}=\sigma_{\varepsilon}=1$ for both input and output variables, and, for simulating outliers, we set $\sigma_{\mathbf{x}}=\sigma_{\varepsilon}=3$. Moreover, for each of the true coefficients of the underlying model, we assumed  $\beta \sim U(-5,5)$ for normal data points and $\beta \sim U(10,15)$ for outliers. Similarly to \cite{Deldossi2022}, the outliers are introduced in the data stream in the form of isolated covariate and concept shifts. That is, an anomalous data point is a point for which we have both a larger variation in the input space and a different relationship with the corresponding response variable. In the simulated scenarios, outliers are randomly distributed in the data stream according to a pre-defined percentage describing the contamination level of the environment. The performance of the models is expressed, in predictive terms, by the root mean squared error (RMSE) of the predictions on a separate test set, only composed of normal observations. This is coherent with the objective of trying to understand the true underlying relationship between predictors and response, and not the erroneous one that could be derived from the outliers.

The effectiveness of the proposed approach is evaluated by comparing the learning curves reporting the average RMSE values for each learning step, which are obtained using 1000 simulations for each scenario. A learning step indicates the acquisition of a new labeled observation and its inclusion in the training set. Hence, at each step, we are comparing models that are trained using the same number of labeled examples. We set the number of process variables equal to 20, the budget constraint $B$ equal to 50, and the warm-up length m to 500. The warm-up length indicates the number of unlabeled observations that are used to estimate the covariance matrix $\mathbf{\Sigma}$ that is used for pre-whitening the observations. With regards to the sampling rate, we used $\alpha=5\%$ for all the sampling strategies and $c=5\%$ for the protection cut-off value used by the bounded CDO algorithm. We selected $5\%$ as it is a commonly employed value, especially when no previous specific knowledge is available.

\subsection{No outliers}
\begin{figure}[h]
  \centering
  \includegraphics[width=.7\linewidth]{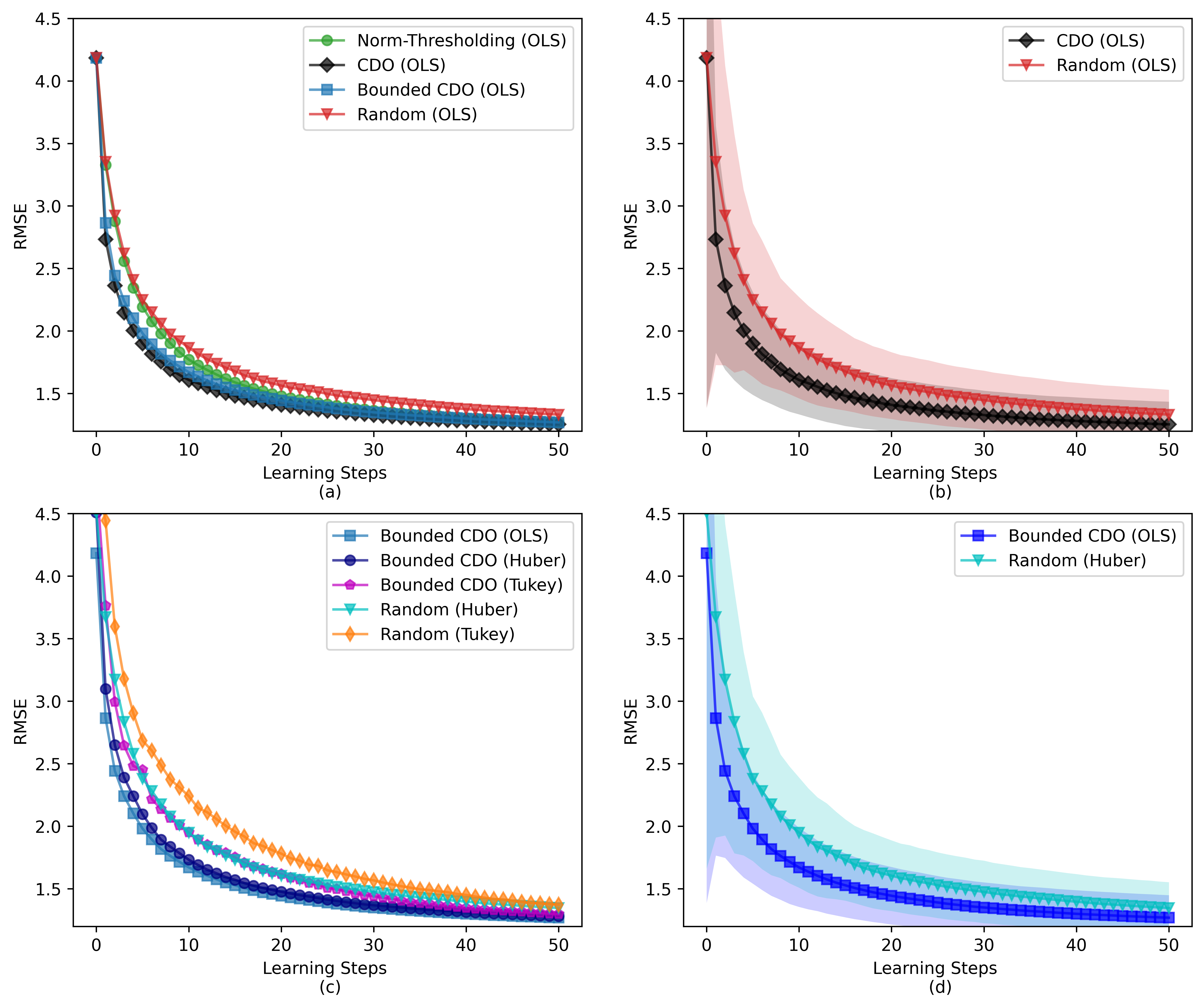}
  \caption{Comparing query strategies in the absence of outliers: results from 1000 simulations. Plots (b) and (d) offer a closer view on the two best strategies from plots (a) and (c), respectively, with shaded regions indicating the standard deviation across the simulations.}
  \label{fig:noout}
\end{figure}

We first evaluated the query strategies to assess their performance in the absence of outliers. Consistently with the findings reported in \cite{SBAL}, our results in Figure \ref{fig:noout} indicate that the standard CDO algorithm performs best when there are no outliers in the data stream. The use of robust estimators does not provide any added value in this scenario. Both the Huber and Tukey estimators are unable to outperform the bounded CDO strategy with the OLS model, which in turn is only marginally worse than the standard CDO. In Figure \ref{fig:noout} , plots (a) and (b) represent the strategies that rely on the OLS models, while plots (c) and (d) show the strategies that use robust models, with the bounded CDO based on OLS included for comparison.

\subsection{0.275\% outliers}
\begin{figure}[h]
  \centering
  \includegraphics[width=.7\linewidth]{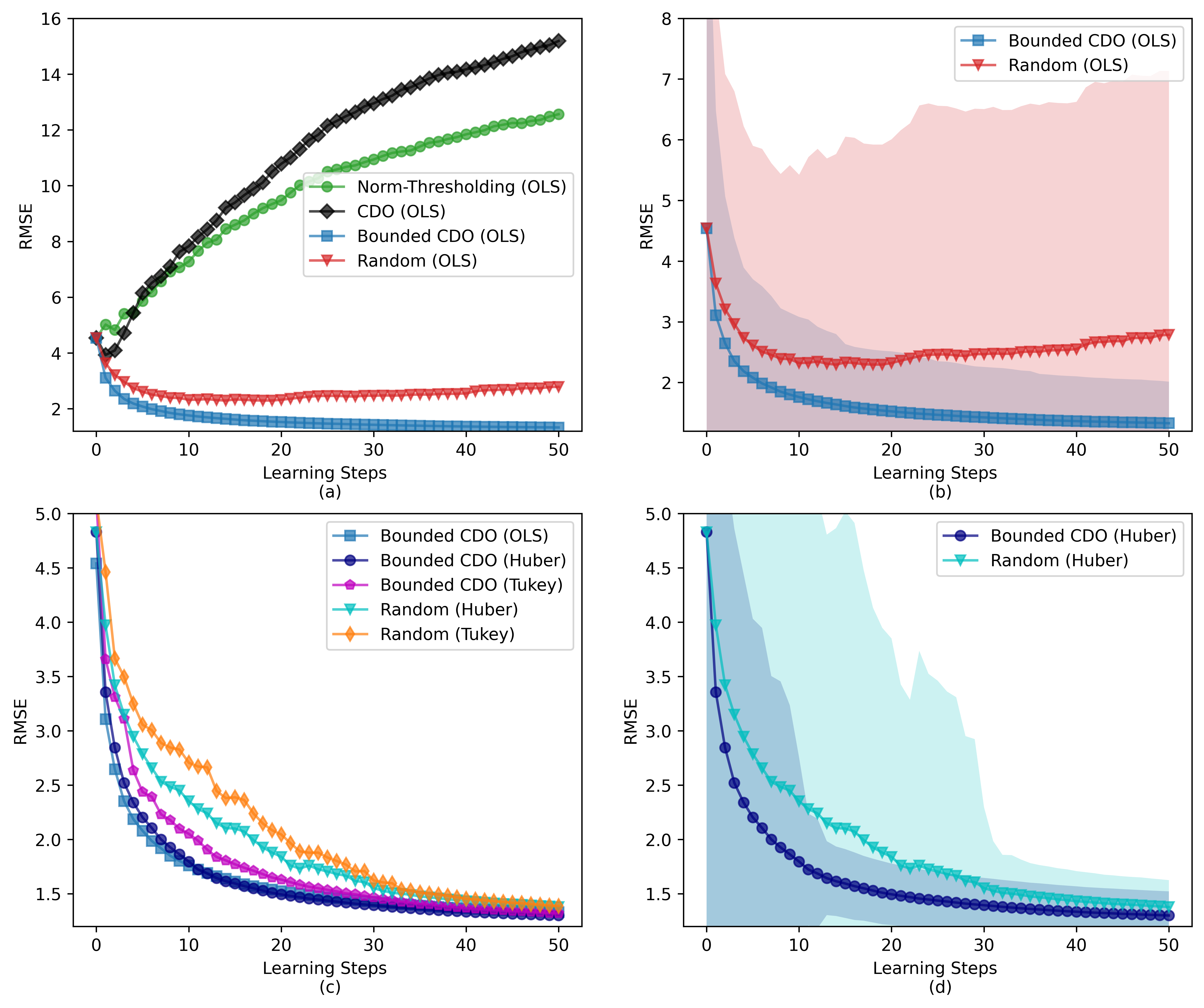}
  \caption{Comparing query strategies with 0.275\% outliers (1000 simulations). Plots (b) and (d) offer a closer view on the two best strategies from plots (a) and (c), respectively, with shaded regions indicating the standard deviation across the simulations.}
  \label{fig:0275clean}
\end{figure}

The second scenario depicts a circumstance where only a modest fraction of the data stream is represented by outliers. We can see from plot (a) of Figure \ref{fig:0275clean} how the performance of the norm-thresholding and the CDO algorithm is dramatically worsened, as they are both prone to sample outliers. The random strategy seems to be a better option and the bounded CDO strategy offers the best results. In plots (c) and (d) of the same figure, we can see the comparison with the results obtained from the robust estimators. In this scenario, using a robust estimator does not seem to offer a significant improvement over the bounded CDO strategy based on OLS. Indeed, the learning curves obtained with the bounded strategy employing the OLS estimator and the Huber estimator are very similar. 

\subsection{1\% outliers}
\begin{figure}[h]
  \centering
  \includegraphics[width=.7\linewidth]{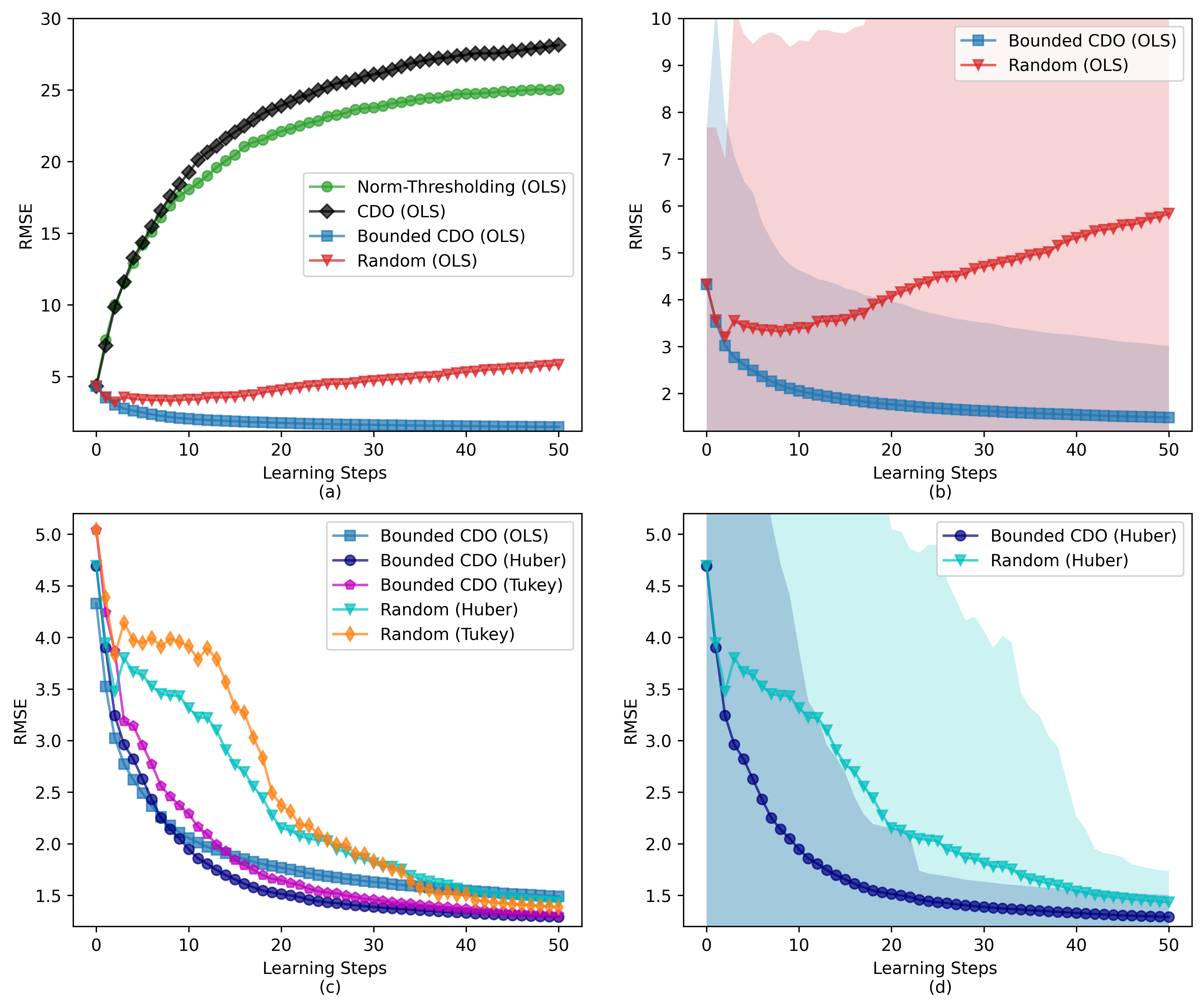}
  \caption{Comparing query strategies with 1\% outliers (1000 simulations): results from 1000 simulations. Plots (b) and (d) offer a closer view on the two best strategies from plots (a) and (c), respectively, with shaded regions indicating the standard deviation across the simulations.}
  \label{fig:001clean}
\end{figure}

The third scenario reports a worse situation, where the process is affected by a large number of outliers, i.e. $1\%$ of the total number of observations. The results in Figure \ref{fig:001clean} are similar to the ones from the previous scenario, with the exception that now the gap between bounded CDO and random sampling is much wider. This should be due to the fact that uniformly sampling observations with $\alpha=5\%$ would most certainly lead to the inclusion of a greater number of outliers in the training set.

\noindent
As per the robust estimators shown in the plots (c) and (d) of Figure \ref{fig:001clean}, it is possible to see how the use of robust estimators now offers an evident value-added, also when compared to the OLS-based bounded CDO. While the learning curves are more or less overlapping in the first five learning steps, the models fitted using the Huber and Tukey losses are yielding a lower prediction error in the remaining steps.

\subsection{5\% outliers}
\begin{figure}[H]
  \centering
  \includegraphics[width=.7\linewidth]{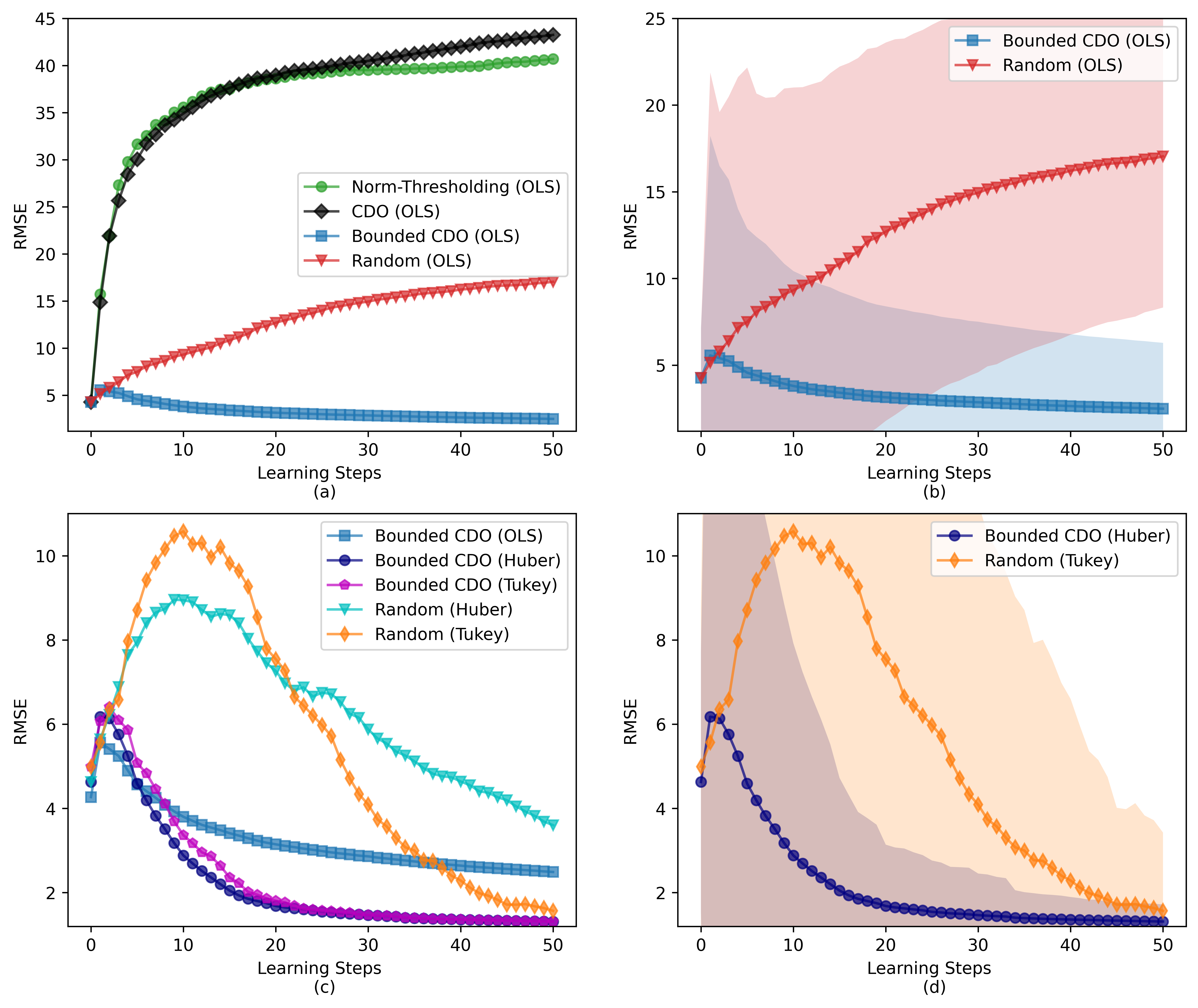}
  \caption{Comparing query strategies with 5\% outliers (1000 simulations): results from 1000 simulations. Plots (b) and (d) offer a closer view on the two best strategies from plots (a) and (c), respectively, with shaded regions indicating the standard deviation across the simulations.}
  \label{fig:005clean}
\end{figure}

The final scenario simulates a pathological case, where 5\% of the observations from the data stream are outliers. The results from the third scenario are exacerbated here. In the case of the OLS estimators, the bounded CDO is still the best strategy, being the only one with a descending learning curve (plots (a) and (b) of Figure \ref{fig:005clean}). Instead, from the plots (c) and (d) of Figure 5 we can see how the robust estimators are able to improve the results obtained with the bounded CDO strategy. In this circumstance, there is not a clear distinction between the Huber and the Tukey models.

%% file: 5_discussion.tex
The experiments presented in this study aimed to evaluate the performance of different active learning strategies in the presence of outliers in a data stream. The results showed that the standard CDO algorithm performed best in the absence of outliers, while the bounded CDO strategy coupled with OLS and robust estimators provided better results when outliers were present. In scenarios where an initial training set free from outliers is available and only a modest fraction of the data stream is represented by outliers, the bounded CDO strategy employing an OLS estimator seems to be the better option. Conversely, in the case of a larger contamination level, sampling strategies based on robust estimators yield the best results. When using robust estimators, for our datasets we did not find solid evidence that using a weighted prediction variance is an advantage. Another interesting observation is that, in the presence of outliers, the standard OLS methods (random, norm-thresholding, and CDO) never converge to the results obtained with the robust query strategies. This is because they tend to accumulate outliers in the training set, which degrade the predictive performance as the model is not allowed to forget old or redundant data. The findings from this study have important consequences for practical applications of active learning strategies, especially in contexts where the data stream is contaminated by outliers. The results suggest that the choice of the active learning strategy should depend on the level of contamination of the data stream. When the data stream is free from outliers, the standard CDO is a good strategy. However, even when a modest fraction of the observations is corrupted, bounding the search area of the active learning algorithm or using robust estimators might be necessary. Overall, this study provides valuable insights into the performance of active learning strategies in the presence of outliers and can inform the development of more effective approaches for real-world applications. However, it is worth noting that the simulations were based on specific assumptions about the data generation process and may not fully capture the complexity of real-world data streams. Further research is needed to validate these findings on real-world datasets and to investigate the generalizability of the proposed approach.

%% file: 6_conclusion.tex
In many real-world problems, data is only available in an unlabeled form, and acquiring the labels is often an expensive and time-consuming task. In these circumstances, active learning is able to reduce the computational burden required to achieve compelling predictive performance by selecting the most informative data points to query. In this paper, we analyze the online active learning framework when the data stream is corrupted by the presence of outliers. In general, we show how the presence of outliers dramatically worsens the performance of the currently proposed methods for active linear regression. To tackle this issue, we propose a modification of the CDO algorithm that filters the outliers, while still focusing on the most promising observations based on the concepts of D-optimality and prediction variance. The analysis shows how this solution is sufficient to make the CDO strategy robust to a modest presence of outliers. When the percentage of outliers in the data stream is higher, the best results are obtained by coupling the bounded CDO strategy with a robust estimator. In general, the proposed approaches can effectively solve the problem of outliers contaminating the data stream, without adding computational complexity compared to the original CDO strategy.

%% file: 7_appendix.tex
\section{Stopping criterion} \label{app:stop}

\begin{figure}[h]
  \centering
  \includegraphics[width=.7\linewidth]{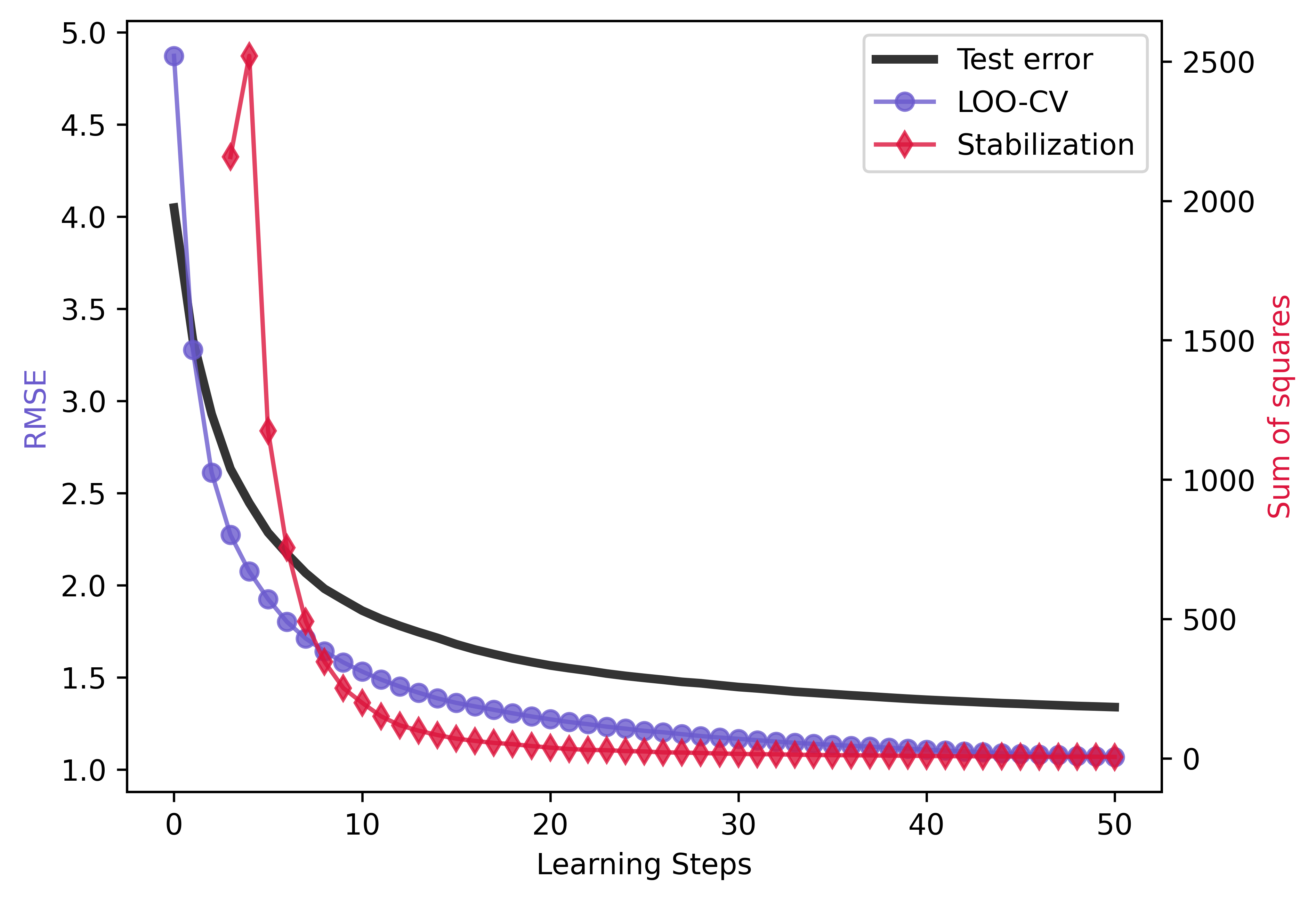}
  \caption{Approximating the learning curve: random sampling with no outliers (1000 simulations). The left axis reports the RMSE value for the curves related to the test error and the LOO-CV. The right axis shows the average sum of squares related to the stabilization score.}
  \label{fig:SC}
\end{figure}

In real-world applications of active learning, if we do not have an explicit operational budget on the number of experiments that can be run, it can be challenging to determine when to stop collecting new labels due to the unavailability of the true learning curves. To address this problem, it is beneficial to approximate the learning curve using proxy measures. In this study, we investigate the use of two proxy measures. Firstly, we propose monitoring the slope of the stabilization score, drawing inspiration from the stabilizing predictions \cite{Bloodgood2009} and validation set agreement \cite{approxLC} methods employed in classification. In the regression framework, we calculate the stabilization of predictions by averaging the sum of squares of the differences between the predictions of the $w$ most recent pairs of models. Similarly to \cite{Bloodgood2009}, we utilize a window size of 3 ($w = 3$). The values being compared are the predicted values of the calibration set $\textbf{C}$, obtained through successive models. As the examples in $\textbf{C}$ are not used in the annotation process, this curve is solely influenced by the impact of selected and labeled examples on training new models. Essentially, this curve monitors when the predictions from models trained with newly included observations start producing highly similar results. The stopping rule can then be determined through visual inspection of the curve, by setting a tolerance for the sum of squares not improving or approaching zero, or by applying a hypothesis testing procedure. Another performance-based metric we consider is the leave-one-out cross-validation (LOO-CV) score obtained by the model on the currently available labeled observations. While this technique relies on ground-truth labels and may appear advantageous, it may not be the optimal choice if the collected training set is biased or does not accurately represent the real data distribution \cite{Farquhar2021}. On the other hand, the stabilization score, despite not relying on real labels, could be more reliable if the calibration set $\textbf{C}$ follows the population distribution. Figure \ref{fig:SC} demonstrates the effectiveness of the two proposed methods in approximating the true test error curve, offering valuable insights for determining when to halt the active learning routine.

\section{Weighted prediction variance}\label{app:upv}

\begin{figure}[H]
  \centering
  \includegraphics[width=.7\linewidth]{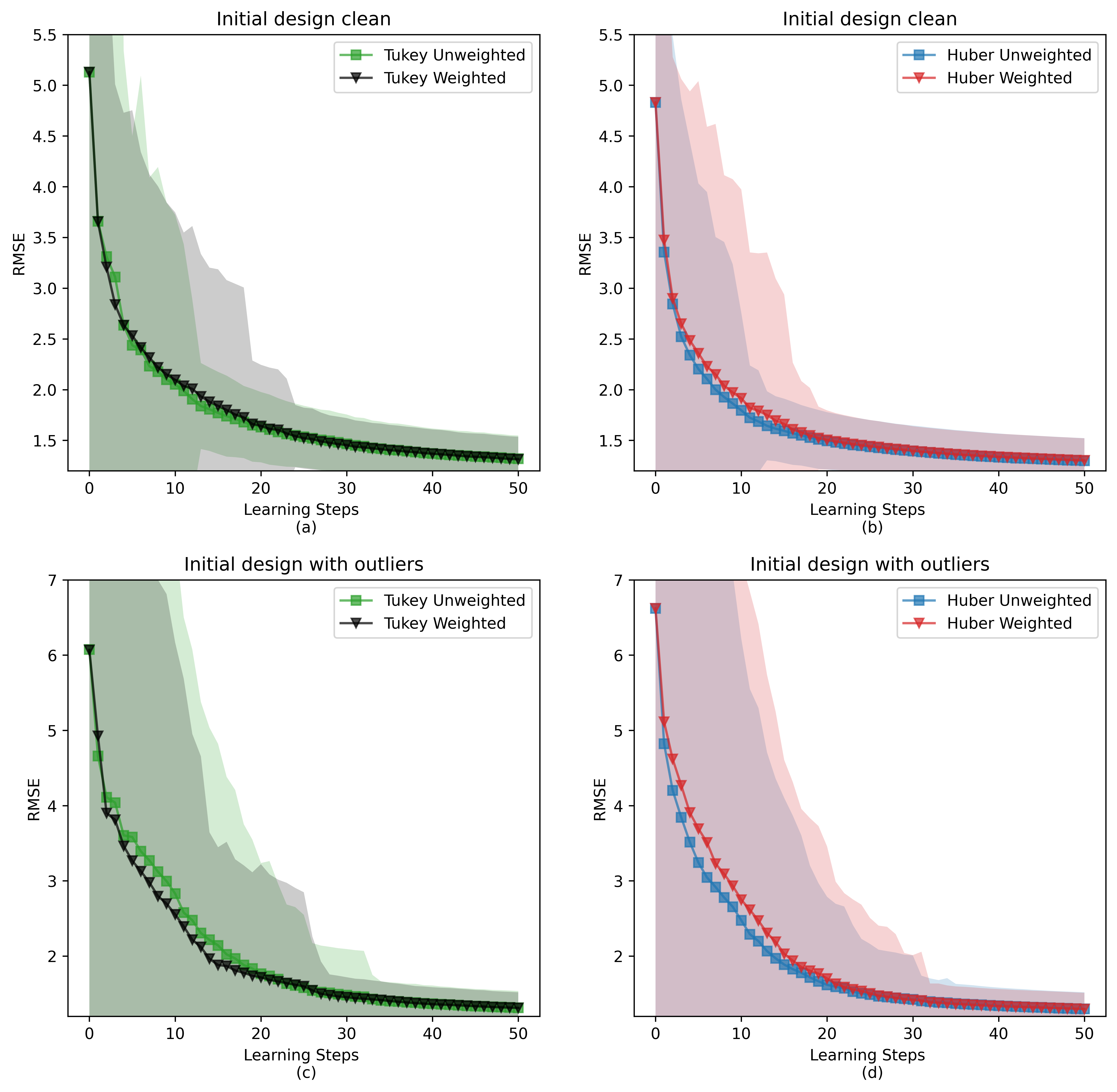}
  \caption{Comparing UPV and $\text{UPV}_w$ in the scenario with 0.275\% outliers (1000 simulations).}
  \label{fig:upv0275}
\end{figure}

\begin{figure}[H]
  \centering
  \includegraphics[width=.7\linewidth]{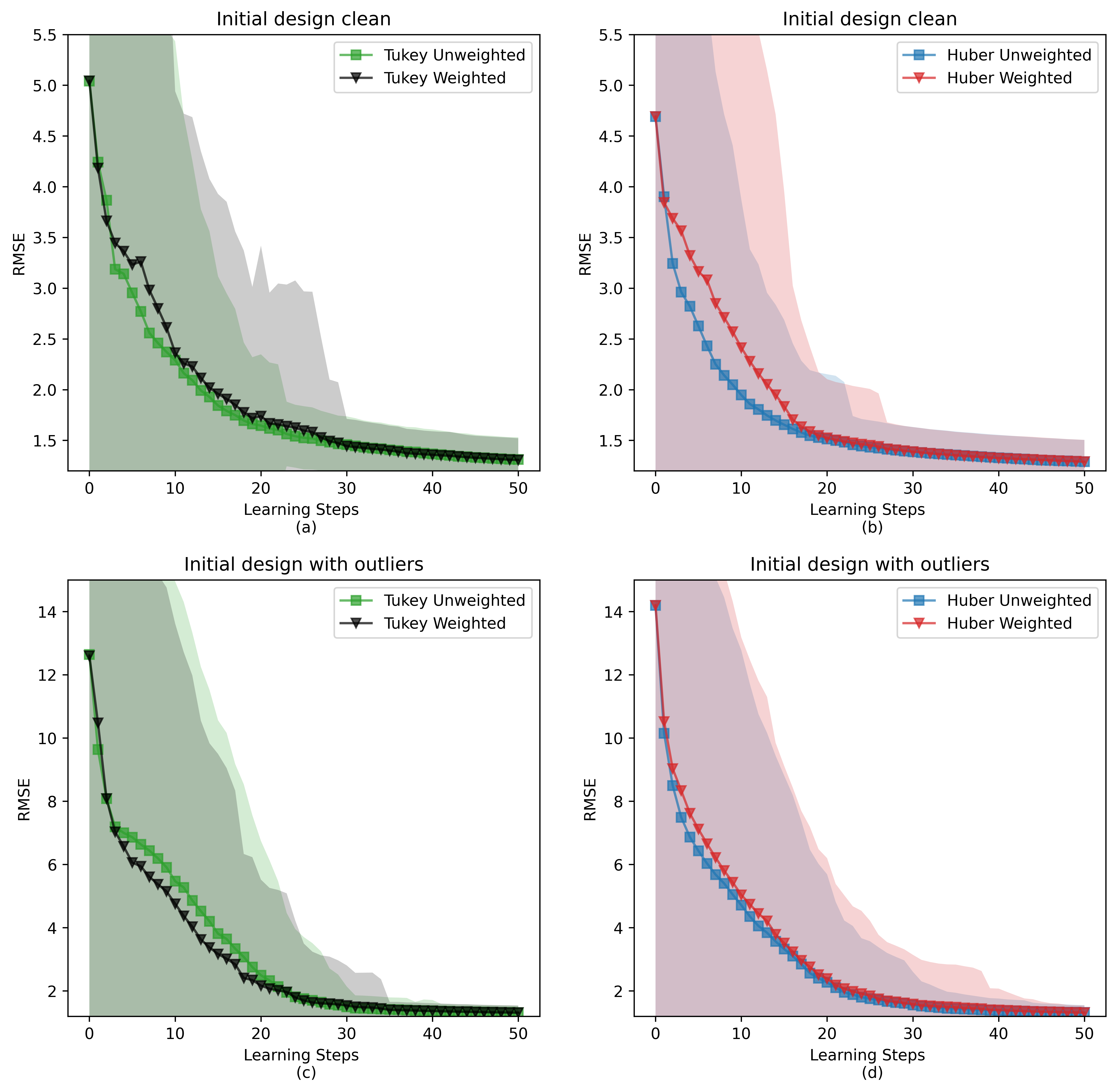}
  \caption{Comparing UPV and $\text{UPV}_w$ in the scenario with 1\% outliers (1000 simulations).}
  \label{fig:upv1}
\end{figure}

\begin{figure}[H]
  \centering
  \includegraphics[width=.7\linewidth]{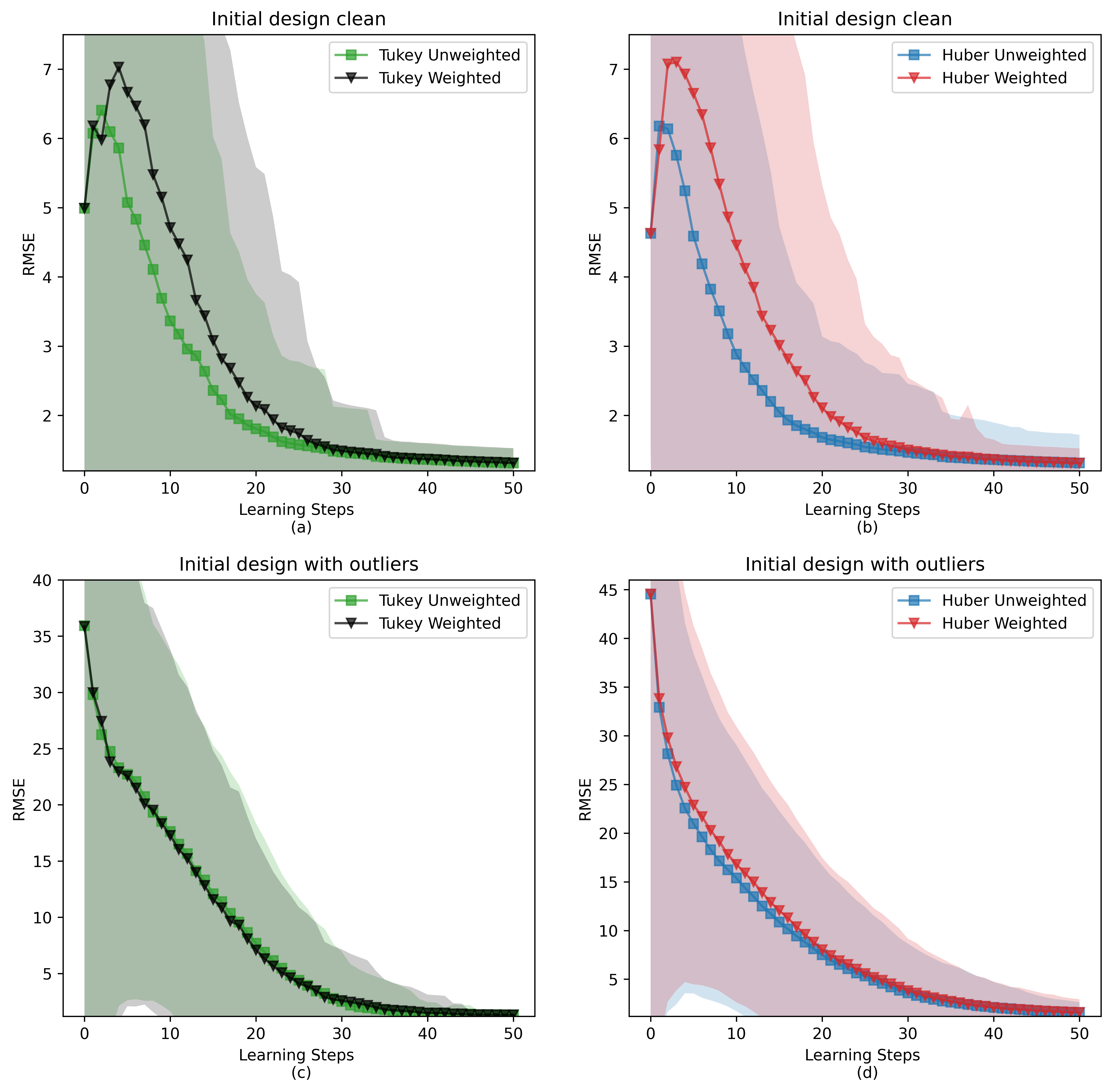}
  \caption{Comparing UPV and $\text{UPV}_w$ in the scenario with 5\% outliers (1000 simulations).}
  \label{fig:upv5}
\end{figure}

In this section, we examine the impact of switching from the standard UPV to its weighted version $\text{UPV}_w$ on the learning curves of the robust bounded CDO strategies. While it may seem reasonable to use a weighted prediction variance from a theoretical standpoint, we found little compelling evidence that it improves the performance (Figures \ref{fig:upv0275}-\ref{fig:upv5}). In fact, we observed that using the $\text{UPV}_w$ actually worsens results when the initial design is free from outliers. This could be because the robust models mistakenly identify some observations as outliers, resulting in $\mathbf{W} \neq \mathbf{I}_{k}$.

\section{Presence of outliers in the initial design}\label{app:contaminated}

\begin{figure}[H]
  \centering
  \includegraphics[width=.7\linewidth]{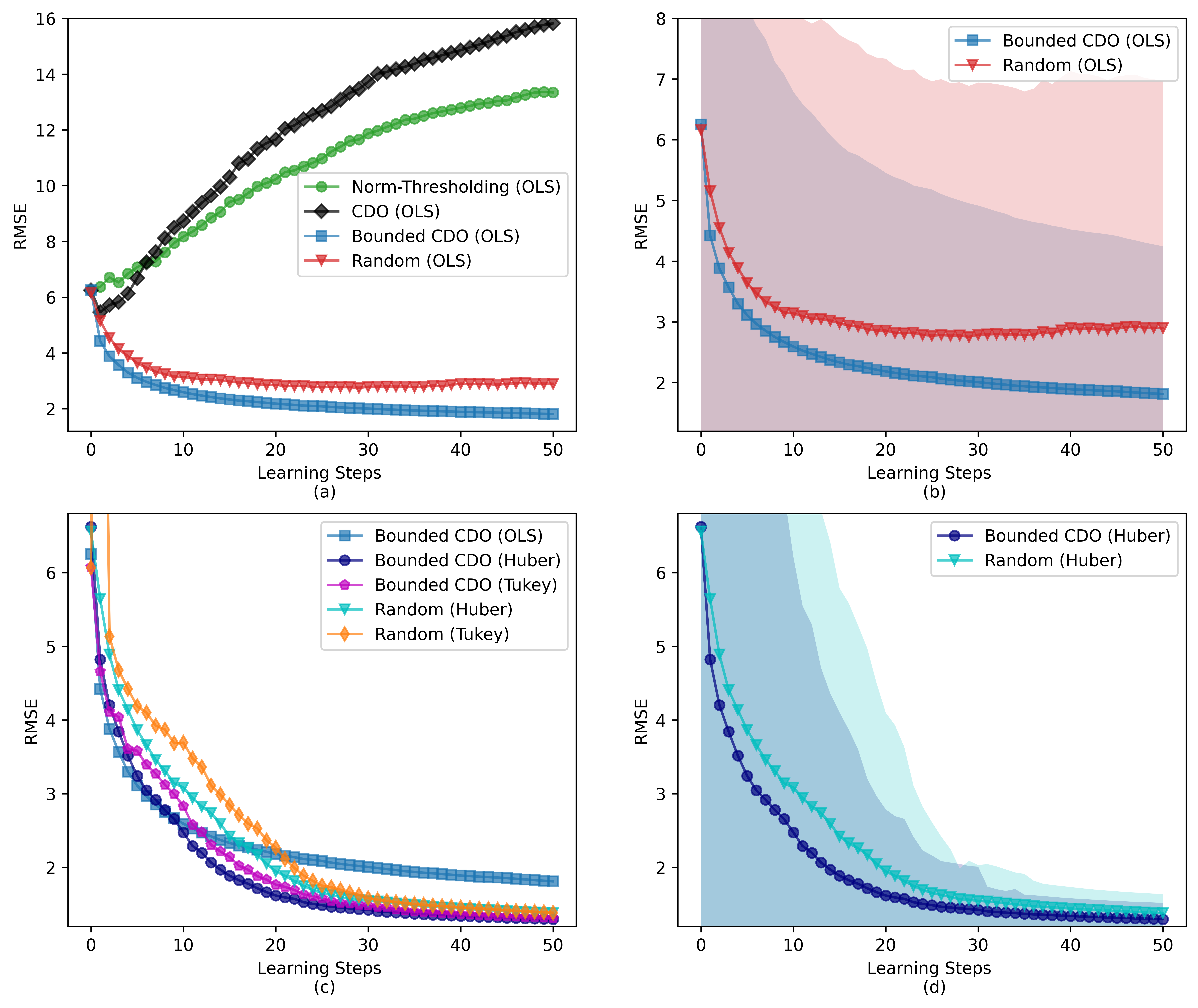}
  \caption{Comparing query strategies with 0.275\% outliers (1000 simulations). Plots (b) and (d) offer a closer view on the two best strategies from plots (a) and (c), respectively, with shaded regions indicating the standard deviation across the simulations.}
  \label{fig:0275dirty}
\end{figure}

\begin{figure}[H]
  \centering
  \includegraphics[width=.7\linewidth]{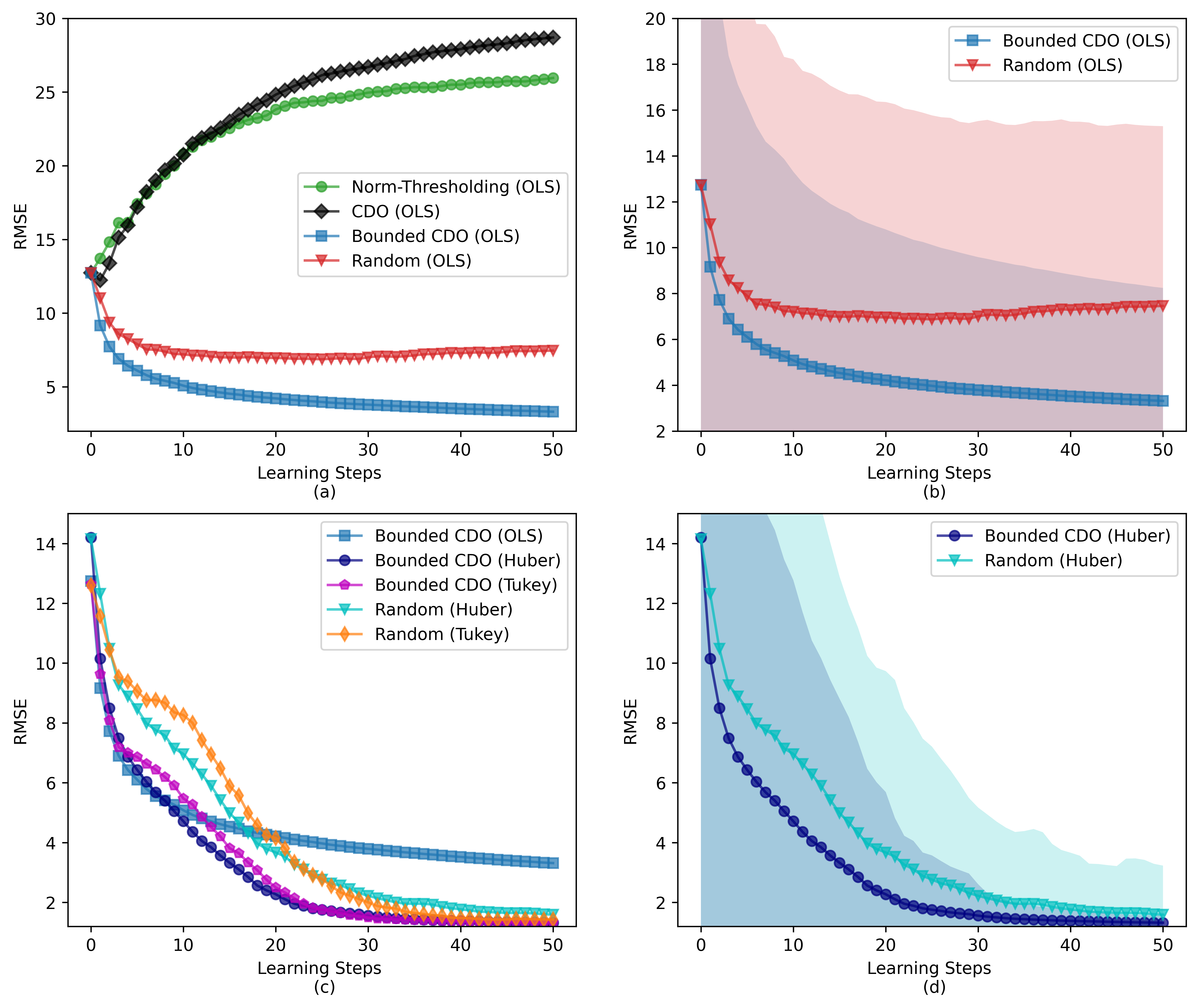}
  \caption{Comparing query strategies with 1\% outliers (1000 simulations): results from 1000 simulations. Plots (b) and (d) offer a closer view on the two best strategies from plots (a) and (c), respectively, with shaded regions indicating the standard deviation across the simulations.}
  \label{fig:001dirty}
\end{figure}

\begin{figure}[H]
  \centering
  \includegraphics[width=.7\linewidth]{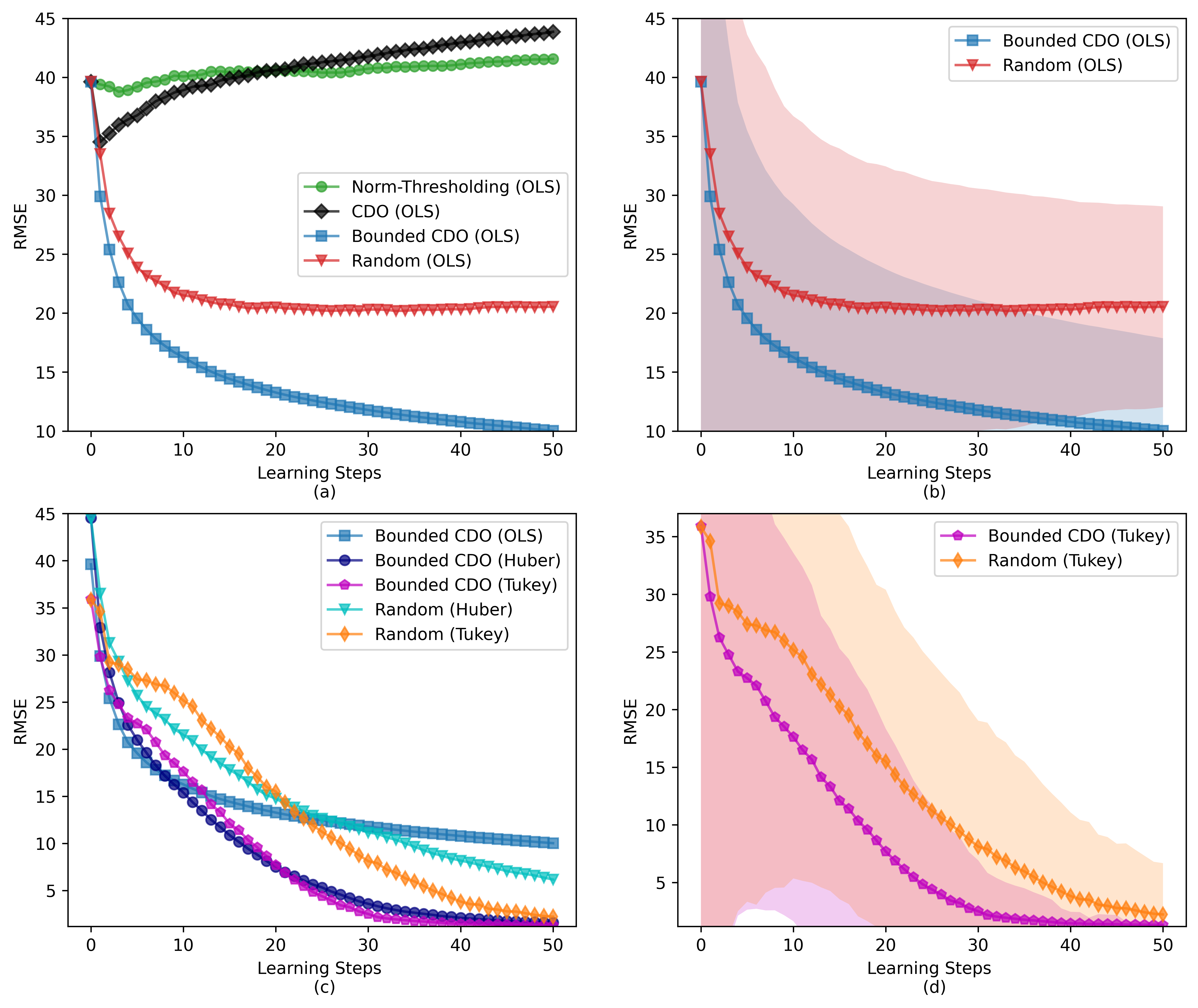}
  \caption{Comparing query strategies with 5\% outliers (1000 simulations): results from 1000 simulations. Plots (b) and (d) offer a closer view on the two best strategies from plots (a) and (c), respectively, with shaded regions indicating the standard deviation across the simulations.}
  \label{fig:005dirty}
\end{figure}

In Figures \ref{fig:0275dirty}-\ref{fig:005dirty}, we investigate the impact of removing the assumption that the initial design is free from outliers on the sampling strategies. Despite the small size of the initial design when $p=20$, we observed several notable behaviors. One of the most noticeable differences is that the learning curves start with higher errors, as there are outliers forcibly included in the data. However, over time, the learning curves of the robust strategies are able to converge to satisfactory predictive performance as they can minimize the impact of these observations on the model training. In contrast, the OLS-based bounded CDO performs significantly worse in this scenario. This is because estimating the cutoff value $\Gamma_2$ using a contaminated set does not provide adequate protection against the inclusion of outliers in the design.

